\documentclass[11pt,a4paper]{article}

\usepackage{times,latexsym}
\usepackage{url}
\usepackage[T1]{fontenc}
\usepackage{booktabs}
\usepackage{amsmath}
\usepackage{amssymb}
\usepackage{tikz}

\newif\ifpreprint
\preprinttrue
\ifpreprint
  \usepackage[acceptedWithA]{tacl2021v1}
\else
  \usepackage[]{tacl2021v1}
\fi
\usepackage{float}

\definecolor{otc}{RGB}{31,119,180}
\definecolor{seh}{RGB}{214,39,40}
\definecolor{cons}{RGB}{44,160,44}
\definecolor{str}{RGB}{148,103,189}
\definecolor{hdn}{RGB}{123,79,110}
\definecolor{fce}{RGB}{129,100,42}
\definecolor{brg}{RGB}{96,131,116}

\title{Beyond Independent Labels: Schwartz-Geometry Decoding for \\ Human Value Detection}

\ifpreprint
  \author{V\'ictor Yeste$^{1,2,*}$ \and Paolo Rosso$^{1,3}$
    \\ \ \\ \mdseries
    $^1$PRHLT Research Center, Universitat Polit\`ecnica de Val\`encia, Spain
    \\
    $^2$School of Science, Engineering and Design, Universidad Europea de Valencia, Spain
    \\
    $^3$Valencian Graduate School and Research Network of Artificial Intelligence (ValgrAI)
    \\ \ \\
    $^{*}$Corresponding author: \texttt{vicyesmo@upv.es}}
\else
  \author{}
\fi
\date{}

\begin{document}
\maketitle

\begin{abstract}
Human value detection is commonly formulated as sentence-level multi-label
classification over the 19 refined Schwartz values, typically predicted as
independent labels. Schwartz theory, however, describes them as a circular
motivational continuum, in which adjacent values are compatible and opposing
values are in tension. We ask whether this structure can be operationalized as an
explicit output-space geometry and used as a soft bias rather than a hard
constraint. On a DeBERTa-v3-base classifier, we compare two ways of injecting it:
training-time geometry-aware objectives and a post-hoc Schwartz-aware energy
decoder that scores whole label sets jointly. Across five seeds, training-time
geometry gives only limited gains---no larger for the true continuum than for a
random ordering---whereas the decoder makes label sets more coherent with the
continuum---on theory-aware coherence metrics we introduce---at no cost to
Macro-F1 or Micro-F1 (held fixed by its selection rule).
The gain is specific to the true Schwartz ordering: it does not appear for a
random permutation or an empirical co-occurrence graph through the identical
decoder. A bounded Qwen2.5-72B-Instruct diagnostic shows that supplying the
continuum at inference shifts behavior but does not match supervised structured
prediction. Theory-aware decoding thus offers a lightweight, controllable way to
make value detection faithful to its label space.
\end{abstract}

\section{Introduction}

Human values underlie moral, social, political, and cultural language, and
detecting them in text supports work across NLP and computational social
science. The task is commonly posed as sentence-level multi-label
classification: given a sentence, predict which refined human values it
expresses \citep{kiesel2023semeval,mirzakhmedova2024touche}. The dominant
modeling approach treats the values as independent labels. This is convenient
but theoretically incomplete: the refined Schwartz theory defines the values not
as independent categories but as a circular motivational continuum, in which
neighboring values are compatible and values on opposite arcs are in tension
\citep{schwartz2012refining,schwartz2017refined,schwartzcieciuch2022measuring}.

We ask whether this theory can be made operational in value detection without
overconstraining it. The goal is not a hard rule that opposite values can never
co-occur---real texts express conflict, compromise, and trade-offs
\citep{schwartz2017tradeoffs,skimina2018realtime}---but a soft inductive bias:
can a model preserve predictive performance while producing label sets that are
more coherent with the Schwartz continuum? The question is timely. Recent work
on the same task shows that strong flat encoders are hard to beat
\citep{ma2023pai,oskuee2023tmscanlon,yeste2026sentence}, while hard architectural
uses of the theory, such as presence gates or higher-order hierarchies, can
introduce recall bottlenecks or error propagation \citep{yeste2026hierarchical}.
It is also sharpened by instruction-tuned LLMs, which can be prompted with the
theory directly \citep{sun2024finetuning,zhu2025eavit}, raising the question of
whether supervised structure is still needed.

We encode the 19 refined values as a circular output-space geometry---an angular
position per value and a circular distance matrix---and use it in two ways: a
training-time penalty and a post-hoc structured decoder over a DeBERTa-v3-base
classifier. Across five seeds, training-time geometry yields only limited,
non-theory-specific gains, whereas the Schwartz decoder improves theory-aware
coherence at no cost to F1, and only for the true continuum---not a random
permutation or an empirical co-occurrence graph. A bounded 72B LLM diagnostic shows
that prompting the theory shifts behavior but does not match the supervised decoder.

We make four contributions: (i) we formulate sentence-level detection of the 19
refined Schwartz human values as a circular output-space geometry, with a family of
theory-aware coherence metrics derived from it; (ii) we compare
training-time geometry-aware objectives with a post-hoc Schwartz-aware energy
decoder on the same classifier, finding that only the decoder yields label sets
more coherent with the continuum; (iii) we isolate the role of the true geometry
through direct random and empirical control comparisons with paired significance
tests; and (iv) we add a bounded LLM diagnostic testing whether prompted theory can
replace supervised structured prediction. By \emph{coherence} we mean a concrete,
measurable property of the predicted label set: fewer false positives on the far
side of the circle and confusions concentrated among nearby values rather than
opposing ones (operationalized in Section~\ref{sec:setup}). Together, the results
support theory-aware decoding as a lightweight, controllable way to make value
detection more faithful to the psychological structure of its label space.

\section{Related Work}

\paragraph{Human value detection.}
Identifying the human values behind arguments was introduced by
\citet{kiesel2022identifying} and scaled into the ValueEval shared task and an
extended benchmark family \citep{kiesel2023semeval,mirzakhmedova2024touche}.
Rooted in argument mining \citep{lawrence2019argument}, the setting also supports
downstream argumentation and deliberation analysis \citep{plenz2024pakt}. More
broadly, detecting values and morality in text is a central goal of computational
social science, as in work grounded in Moral Foundations Theory, its annotated
corpora, and its links to language models
\citep{graham2009liberals,hoover2020moral,zangari2024survey}. These moral and
value signals also support downstream tasks such as hate-speech detection and
identifying violent radicalization \citep{vargas2026hate,yin2026moralization}.
Beyond argument mining, recent work measures the (often subjective) expression
of basic Schwartz human values directly in social-media posts
\citep{epstein2026whose}, underscoring that value attributions are annotator- and
context-dependent.
Strong shared-task
systems rely on transformer encoders---often DeBERTa---together with
class-imbalance handling, threshold tuning, and ensembling
\citep{ma2023pai,oskuee2023tmscanlon,kandru2023tenzin,tsunokake2023hitachi,aydin2024edward}.
Recent work on sentence-level detection of the refined values further reports
that such direct encoders and calibration \citep{guo2017calibration} are strong,
while presence gates and
higher-order hierarchies can introduce recall bottlenecks or error propagation
\citep{yeste2026hierarchical, yeste2026sentence}. Across this line, however, the
values are typically predicted as independent labels or organized by a hard
hierarchy, and the motivational geometry of the label space is left implicit. We
instead make that geometry an explicit, soft component of the output space.

\paragraph{Schwartz values and motivational structure.}
The refined Schwartz theory, building on the original circular value model
\citep{schwartz1992universals}, arranges 19 basic values on a circular
motivational continuum in which adjacent values share compatible goals and
opposing values express tension \citep{schwartz2012refining,schwartz2017refined}---a
structure with broad cross-cultural psychometric support and predictive links to
behavior \citep{schwartzcieciuch2022measuring,cieciuch2013applying,bardi2003values}.
This makes the label space theoretically structured rather than arbitrary. We use
the continuum as an inductive bias and evaluation lens, not as a hard constraint
on what a text may express, since real arguments can voice value tensions and
trade-offs \citep{schwartz2017tradeoffs,skimina2018realtime}. Circular, continuous
structure of this kind is not unique to values: affective constructs are organized
as a circumplex \citep{russell1980circumplex} or an emotion wheel
\citep{plutchik1980general}, and such structured label spaces are increasingly
modeled in NLP \citep{demszky2020goemotions}, which motivates treating the
continuum as an output-space geometry rather than as unordered classes.

\paragraph{Structured multi-label prediction.}
Multi-label learning has long modeled label dependence \citep{tsoumakas2007multi}---through
classifier chains \citep{read2011classifier}, neural sequence-generation decoders
\citep{yang2018sgm}, label-correlation and embedding methods, and graph-based
structures \citep{zhang2014review,huang2024label,tarekegn2024deep}. A
complementary line casts prediction as structured inference or energy
minimization over label configurations, from conditional random fields and
collective classification \citep{lafferty2001conditional,ghamrawi2005collective}
to structured prediction energy networks \citep{belanger2016structured}; our
decoder shares this view but fixes the pairwise term from theory rather than
learning it. In these approaches the
dependencies are typically learned from training co-occurrence or label
semantics. Our work keeps a strong local classifier and adjusts only the final
prediction with a structured decoding objective; the distinguishing feature is
that the structure is derived from a psychological theory and tested directly
against random and empirical control geometries.

\paragraph{LLMs for value classification.}
Prompted large language models are increasingly applied to value and moral
classification---comparing prompting with fine-tuning, value identification and
annotation
\citep{sun2024finetuning,zhu2025eavit,milkova2026measuring,yeste2026context,delacruz2025valuelens},
probing Schwartz value priorities \citep{segerer2025cultural}, assessing moral
abilities against human labels \citep{bulla2025moral}, analyzing values in
real-world interactions \citep{huang2025values}, and ensembling detectors
\citep{rodrigues2024beyond}. Rather than benchmarking many models, we include one
bounded diagnostic asking whether supplying the continuum at inference time can
match incorporating it through supervised training and decoding.

\section{Task and Schwartz Geometry}
\label{sec:task}

\subsection{Task and Data}

We study sentence-level human value detection as multi-label classification
over the 19 refined Schwartz values. Given a sentence \(x\), a model predicts a
binary vector \(y \in \{0,1\}^{19}\), where \(y_k=1\) indicates that value
\(v_k\) is expressed in \(x\). We use the Touch\'e24-ValueEval data family
\citep{kiesel2023semeval,mirzakhmedova2024touche}, segmented and annotated at
the sentence level following recent work on refined-value detection
\citep{yeste2026sentence}.\footnote{The corpus is distributed under a
restricted Data Usage Agreement via Zenodo (The ValuesML Team,
\emph{Touch\'e24-ValueEval}, 2024,
\url{https://doi.org/10.5281/zenodo.13283288}).}

Each value carries two stance annotations, \emph{attained} and
\emph{constrained}. Because our object of study is the geometry of value
\emph{presence} rather than stance, we collapse the two stances into a single
presence indicator: \(y_k=1\) iff value \(v_k\) is annotated as attained or
constrained, and \(0\) otherwise. This yields a standard binary multi-label
target compatible with sigmoid outputs and threshold-based decoding.

We use the official train/validation/test partition, split by document
(\texttt{Text-ID}) so no document is shared across splits and sentence-level overlap
cannot inflate test performance (Table~\ref{tab:data}). About half of all sentences
express at least one value (50.8--51.5\% across splits), and value-positive sentences
are predominantly single-label (\(\approx\!1.1\) values each); multi-value sentences
are a non-negligible minority (e.g., 901 test sentences). Per-label support is
strongly imbalanced, from \emph{Humility} (0.24\%) to \emph{Security: societal}
(8.6\%; full distribution in Appendix~\ref{app:labels}). All hyperparameters and
thresholds are selected on validation only; the test split is used once.

\begin{table}[t]
\centering
\small
\setlength{\tabcolsep}{4pt}
\begin{tabular}{lrrr}
\toprule
 & Train & Dev & Test \\
\midrule
Documents & 1{,}603 & 523 & 522 \\
Sentences & 44{,}758 & 14{,}904 & 14{,}569 \\
\quad \(\ge\!1\) value & 23{,}062 & 7{,}600 & 7{,}402 \\
\quad (\% of split) & 51.5 & 51.0 & 50.8 \\
\quad \(>\!1\) value & 2{,}640 & 876 & 901 \\
Values / positive sent. & 1.13 & 1.13 & 1.14 \\
\bottomrule
\end{tabular}
\caption{Dataset statistics. Splits are partitioned by document, with no
\texttt{Text-ID} shared across splits. The label space is the 19 refined
Schwartz values in all splits.}
\label{tab:data}
\end{table}

\subsection{Circular Value Geometry}

\begin{figure*}[t]
\centering
\begin{tikzpicture}
  \def\R{1.95}
  \draw[gray!35] (0,0) circle (\R);
  \draw[dashed,gray!60,line width=0.4pt] (99.47:\R) -- (0,0);
  \draw[dashed,gray!60,line width=0.4pt] (33.16:\R) -- (0,0);
  \draw[dashed,gray!60,line width=0.4pt] (-42.63:\R) -- (0,0);
  \draw[dashed,gray!60,line width=0.4pt] (-156.32:\R) -- (0,0);
  \foreach \i/\col in {0/otc,1/otc,2/otc,3/hdn,4/seh,5/seh,6/seh,7/fce,8/cons,9/cons,10/cons,11/cons,12/cons,13/brg,14/str,15/str,16/str,17/str,18/str}{
    \pgfmathsetmacro{\ang}{90-\i*360/19}
    \pgfmathtruncatemacro{\num}{\i+1}
    \node[circle,fill=\col,text=white,font=\tiny\bfseries,inner sep=0pt,minimum size=4.4mm] at (\ang:\R) {\num};
  }
  \node[otc,font=\footnotesize\bfseries,align=center] at (55:2.85) {Openness\\to change};
  \node[seh,font=\footnotesize\bfseries,align=center] at (-12:3.05) {Self-\\enhancement};
  \node[cons,font=\footnotesize\bfseries,align=center] at (-115:2.6) {Conservation};
  \node[str,font=\footnotesize\bfseries,align=center] at (148:3.15) {Self-\\transcendence};
\end{tikzpicture}

\medskip
{\scriptsize
\setlength{\tabcolsep}{10pt}
\begin{tabular}{@{}lll@{}}
\textcolor{otc}{1.\ Self-direction: thought} & \textcolor{fce}{8.\ Face} & \textcolor{str}{15.\ Benevolence: dependability}\\
\textcolor{otc}{2.\ Self-direction: action} & \textcolor{cons}{9.\ Security: personal} & \textcolor{str}{16.\ Benevolence: caring}\\
\textcolor{otc}{3.\ Stimulation} & \textcolor{cons}{10.\ Security: societal} & \textcolor{str}{17.\ Universalism: concern}\\
\textcolor{hdn}{4.\ Hedonism} & \textcolor{cons}{11.\ Tradition} & \textcolor{str}{18.\ Universalism: nature}\\
\textcolor{seh}{5.\ Achievement} & \textcolor{cons}{12.\ Conformity: rules} & \textcolor{str}{19.\ Universalism: tolerance}\\
\textcolor{seh}{6.\ Power: dominance} & \textcolor{cons}{13.\ Conformity: interpersonal} & \\
\textcolor{seh}{7.\ Power: resources} & \textcolor{brg}{14.\ Humility} & \\
\end{tabular}}
\caption{The refined Schwartz value continuum used as the output-space geometry.
The 19 values are placed in canonical order around the circle and colored by
higher-order region; Hedonism (4), Face (8) and Humility (14) bridge adjacent regions. Adjacent
values are motivationally compatible, while values on opposite arcs are in
tension. The four dashed radial lines mark the two bipolar dimensions (openness to
change vs.\ conservation; self-enhancement vs.\ self-transcendence).}
\label{fig:circle}
\end{figure*}
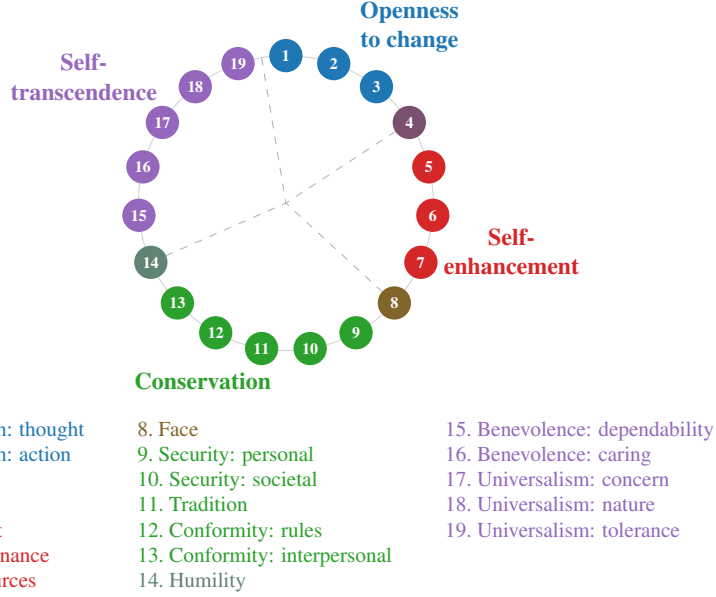

The refined Schwartz theory arranges the 19 values on a circular motivational
continuum: adjacent values share compatible motivational goals, whereas values
on opposite arcs express conflicting motivations
\citep{schwartz2012refining,schwartz2017refined,schwartzcieciuch2022measuring}---a
compatibility-and-conflict structure with confirmatory empirical support
\citep{schwartzboehnke2004evaluating}.
The continuum spans four higher-order regions---\emph{openness to change},
\emph{self-enhancement}, \emph{conservation}, and \emph{self-transcendence}---with
Hedonism, Face, and Humility bridging adjacent regions (Figure~\ref{fig:circle}).
We fix this canonical order as the backbone of an output-space geometry.

We place value \(v_k\) at angle \(\theta_k = 2\pi k/19\), \(k\in\{0,\ldots,18\}\),
on the unit circle, with embedding \(e_k=(\cos\theta_k,\sin\theta_k)\). The
normalized circular distance between two values is the shorter arc between their
positions,
\begin{equation}
\begin{aligned}
d(v_i,v_j) &= \frac{\min\!\big(|\theta_i-\theta_j|,\,2\pi-|\theta_i-\theta_j|\big)}{\pi}\\
&= \frac{2}{19}\,\min\!\big(|i-j|,\,19-|i-j|\big),
\end{aligned}
\label{eq:dist}
\end{equation}
so that \(d=0\) for identical values and \(d\) approaches \(1\) for diametrically
opposed values. Because 19 is odd there is no exact antipode, and the maximum
attainable distance is \(18/19\approx0.95\). The resulting \(19\times19\) matrix
\(D\) is the single quantity from which every geometry-aware term in the paper is
derived.

From \(D\) we read off two relations used later by the objectives, decoder, and
metrics: \textbf{neighbor compatibility}---pairs within two steps on the circle
(\(1\le\min(|i-j|,19-|i-j|)\le2\)), expected to co-occur more plausibly---and
\textbf{opposite tension}---pairs with \(d>0.75\) (at least eight steps apart),
treated as conflicting and requiring stronger evidence to be predicted jointly.
We stress that this geometry is a soft inductive bias, not a hard constraint:
the refined structure has broad cross-cultural psychometric support
\citep{schwartzcieciuch2022measuring}, but real texts can express tension or
compromise, so opposing values are penalized rather than forbidden.

To test whether any benefit is specific to the \emph{true} Schwartz structure
rather than to structure in general, we compare \(D\) against two controls built
with the same machinery. A \textbf{random circular geometry} applies
Equation~\ref{eq:dist} to a seeded random permutation of the 19 values, preserving
the circular form but destroying the theory-derived ordering. An \textbf{empirical
co-occurrence geometry} uses distances \(1-\mathrm{Jaccard}(i,j)\) from
training-split label co-occurrence \citep{huang2024label}, capturing data-derived
dependency rather than motivational theory.

\section{Methods}
\label{sec:methods}

All supervised systems share one architecture: a DeBERTa-v3-base encoder
\citep{he2023debertav3} with a linear head mapping the pooled sentence
representation to 19 logits \(z\), with \(p=\sigma(z)\). They differ only in (i) the
training objective and (ii) an optional post-hoc decoding step; the encoder,
optimization, data protocol, and seeds are held fixed (Section~\ref{sec:setup}), so
differences reflect the objective or decoder, not model capacity. We use the
distance matrix \(D\) and its neighbor and opposite relations throughout. Our main
method is the post-hoc decoder of Section~\ref{sec:decoder}; training-time geometry
(Section~\ref{sec:geotrain}) is an alternative way to inject the same structure. We use three adjectives with distinct scopes: \emph{theory-aware} (faithful to
Schwartz theory by any mechanism), \emph{geometry-aware} (any method using the
distance matrix \(D\), including the controls), and \emph{Schwartz-aware}
(specifically the true continuum).

\subsection{Independent Supervised Baselines}

The primary baseline trains the classifier with binary cross-entropy (BCE) over
the 19 labels, treating them as independent. As an imbalance-aware baseline we
also train with asymmetric loss (ASL) \citep{benbaruch2021asymmetric}, which
adds two terms to BCE: it clips and down-weights easy negative labels through a
focal-style modulation, focusing learning on positives and hard negatives. This
is motivated by the strong label imbalance in the data (Appendix~\ref{app:labels}).
At inference, both baselines convert probabilities to labels with per-label
thresholds \(\tau\) tuned on validation, \(\hat{y}_k = \mathbb{1}[p_k \ge \tau_k]\).

\subsection{Geometry-Aware Training}
\label{sec:geotrain}

Training-time variants inject the geometry directly into the objective, leaving
inference unchanged. \textbf{GeoLoss} adds to a base loss
\(\mathcal{L}_{\text{base}}\in\{\text{BCE},\text{ASL}\}\) a distance-weighted
penalty \(\lambda\,\mathbb{E}_x\big[\sum_{i,j} y_i D_{ij} p_j \big/ \sum_i y_i\big]\),
so that probability mass placed on values far from the gold set on the circle is
penalized in proportion to its circular distance. \textbf{GeoSmooth} instead
replaces the binary targets with geometry-smoothed soft targets
\(\tilde{y}_j = \max_i\, y_i \exp(-D_{ij}^2/\tau)\) (clamped to retain the
original positives) before applying the base loss, giving nearby values a small
amount of soft supervision; GeoSmooth is thus a distance-aware form of label
smoothing \citep{szegedy2016rethinking,pereyra2017regularizing,muller2019does,diaz2019soft}. Both \(\lambda\) and
\(\tau\) are tuned on validation.

The same machinery defines our structure controls by swapping the matrix \(D\):
\textbf{random GeoLoss} uses a seeded random circular permutation, and
\textbf{empirical structure} uses the co-occurrence distances of
Section~\ref{sec:task}. If any structure helps, the random control should help
too; if data-derived dependency suffices, the empirical control should match the
true geometry.

\subsection{Schwartz-Aware Energy Decoder}
\label{sec:decoder}

Our main method keeps the trained classifier fixed and replaces independent
thresholding with a structured decoder applied to its probabilities. Intuitively,
it picks the label set that keeps the labels the classifier already supports,
rewards co-selecting compatible neighbors, and penalizes co-selecting opposing
values---a structured alternative to thresholding each label in isolation. For a
sentence \(x\), the decoder selects the label set that maximizes a structured
score---equivalently, the negative of an energy in the sense of energy-based
models \citep{lecun2006tutorial}---that combines classifier evidence with the
Schwartz geometry,
\begin{equation}
\begin{aligned}
\hat{y} = \operatorname*{arg\,max}_{y \in \mathcal{Y}(x)} \Big[\,
  & \sum_{k} y_k\, u_k(x) \\
  &{}+ \tfrac{1}{2}\sum_{i\neq j} y_i y_j\, W_{ij} \\
  &{}- \gamma\,\big(\textstyle\sum_k y_k - 1\big)_{+} \,\Big].
\end{aligned}
\label{eq:decoder}
\end{equation}
The unary term \(u_k(x)=\sigma^{-1}(p_k(x))-\sigma^{-1}(\tau_k)\) is the
classifier's log-odds margin over the validation-tuned threshold, so it is
positive exactly when \(p_k \ge \tau_k\). The pairwise weights
\(W_{ij}=\alpha N_{ij}-\beta O_{ij}\) combine the neighbor-compatibility matrix
\(N\) (a soft weight that is largest for immediate neighbors and decays to zero
beyond two steps) and the opposite mask \(O\) (one for pairs with
\(D_{ij}>0.75\)): with \(\alpha,\beta\ge 0\), co-selecting neighbors is rewarded
and co-selecting opposing values is penalized. The final term is a cardinality
penalty (\(\gamma\ge 0\), applied beyond the first label) that discourages
over-large sets. When \(\alpha=\beta=\gamma=0\) the maximizer reduces exactly to BCE
thresholding, so the decoder is a strict generalization of the baseline. To keep
the maximization exact and cheap, \(\mathcal{Y}(x)\) is restricted to a small pool
of high-scoring labels (threshold-positive, high-probability, and top-ranked),
capped at a few candidates and a few decoded labels per sentence; exact cut-offs
are in Section~\ref{sec:setup}.

The weights \((\alpha,\beta,\gamma)\) are tuned on validation under a
Pareto criterion: among settings that retain Macro-F1 within a small tolerance
of the best validation Macro-F1, we select the one that minimizes a validation
geometry cost (a label-set coherence measure defined in Section~\ref{sec:setup}). This is, by construction, the reason the decoder improves
theory-aware coherence while preserving F1: F1 preservation is enforced by the
Pareto constraint rather than discovered empirically. The test set is decoded once with the selected
weights. The identical decoder is run with the Schwartz, random, and empirical
geometries (\(N\), \(O\) derived from each), providing the same controls as in
training. Like other structured multi-label methods it adjusts a strong local
classifier's output, but its pairwise structure is theory-derived rather than
learned from data \citep{read2011classifier,zhang2014review}.

\subsection{LLM Diagnostic}

To test whether the theory can instead be supplied at inference time to a large
language model, we include a bounded diagnostic with Qwen2.5-72B-Instruct
\citep{qwen2025technical} under two prompts. Both state the task, the attained/constrained-to-presence
convention, and the 19 value definitions, and require a strict JSON label list
drawn only from the allowed values; they differ only in that the
\emph{continuum} prompt additionally describes the circular ordering and the
expectation that nearby values are compatible while opposing values usually
conflict. Decoding is deterministic (temperature 0). Outputs are parsed against
the exact allowed label set, and we record an invalid-output rate for responses
that cannot be parsed without repair. This isolates theory injected through
prompting from theory injected through training or decoding; full prompts are in
Appendix~\ref{app:prompts}.

\section{Experimental Setup}
\label{sec:setup}

\paragraph{Data and protocol.}
All systems use the sentence-level splits of Section~\ref{sec:task} with the
document-level partition. Hyperparameters and thresholds are selected on
validation; the test split is evaluated once. Each supervised configuration is
run with five seeds (42, 7, 1701, 11, 1984), and we report mean and standard
deviation over seeds.

\paragraph{Backbone and optimization.}
All supervised systems fine-tune \texttt{deberta-v3-base} with AdamW
\citep{loshchilov2019decoupled}. Base hyperparameters are selected once on
validation by grid search over learning rate (\(\{6,7,8,9,10\}\times10^{-6}\)) and
weight decay (\(\{0.10,\ldots,0.20\}\)), yielding \(10^{-5}\) and \(0.15\);
effective batch size (16), sequence length (1024 tokens), and gradient clipping
(\(1.0\)) are fixed. We train up to 30 epochs with early stopping (patience 3) and
tune per-label thresholds \(\tau\) on validation by sweeping \([0,1]\) in steps of
\(0.01\) to maximize each label's F1.

\paragraph{Hyperparameter selection.}
Objective- and decoder-specific hyperparameters are likewise selected on
validation. For ASL we search \(\gamma_{\text{neg}}\in\{2,3,4,5\}\) and
clip\,\(\in\{0,0.03,0.05,0.1\}\) with \(\gamma_{\text{pos}}=0\); for GeoLoss
\(\lambda\in\{0.01,0.05,0.1,0.2\}\); and for GeoSmooth
\(\tau\in\{0.1,0.2,0.5,1.0\}\). The energy decoder uses fixed component
magnitudes (neighbor \(\alpha=0.1\), opposite \(\beta=0.2\), cardinality
\(\gamma=0.02\)) and a two-step neighbor window. The candidate pool for each
sentence is the union of the threshold-positive labels, labels with probability
above half the tuned threshold or above \(0.01\), and the eight highest-probability
labels; it is capped at eight candidates and the decoded set at five labels.
Validation selection under the Pareto rule of Section~\ref{sec:decoder} (retain
validation Macro-F1 within \(99\%\) of thresholding, then minimize the validation
geometry cost) determines which components are active; in our final runs it keeps
the neighbor and opposite terms but sets the cardinality term to \(\gamma=0\).

\paragraph{Metrics.}
We report three standard metrics---Macro-F1, Micro-F1, and threshold-free
Macro-AUPRC---together with a family of theory-aware metrics that we introduce
for this label space (in the spirit of distance-weighted and ordinal error
measures such as the earth mover's distance
\citep{rubner2000earth,hou2016squared} and of hierarchical-classification
evaluation \citep{kosmopoulos2015evaluation}), derived from the circular geometry. Because probabilities and discrete label sets call for different
theory-aware measures, the supervised models and the decoder are scored with
different ones (hence the differing theory-aware columns in
Tables~\ref{tab:supervised-results} and \ref{tab:decoder-main}). For supervised
models, which emit probabilities, we report the
\emph{expected circular error}: the gold-normalized, distance-weighted
probability mass placed away from the gold values (lower is better), i.e.\ the
quantity penalized by GeoLoss. For the structured decoder, which emits label
sets, we report label-set metrics: the \emph{opposite-error rate}, the share of
false-positive labels that are opposite (\(D>0.75\)) to every gold value (lower
is better); the \emph{neighbor-error rate}, the share falling within two steps
of a gold value (near-miss errors); and the \emph{confusion-distance
correlation} between label-pair circular distance and confusion frequency. The
\emph{decoder geometry cost} sums these three---two rates in \([0,1]\) and a
correlation in \([-1,1]\), equally weighted---so a lower value is more coherent.
All three are corpus-level quantities computed once per configuration, so selection
picks a single global \((\alpha,\beta,\gamma)\) under the Macro-F1 constraint, not a
per-sentence rule. As the composite is dimensionally heterogeneous, we report its
components separately (Section~\ref{sec:decoder-results}) and treat the
Schwartz-vs-control contrast, not the absolute cost, as the primary evidence.

\paragraph{Significance testing.}
For decoder and LLM comparisons we use paired, sample-level bootstrap tests
\citep{dror2018hitchhiker} (2{,}000 resamples) on the shared test set, reporting
two-sided \(p\)-values at the \(0.05\) level. This includes the direct Schwartz-vs-control tests on
decoder geometry cost. For the supervised systems
(Table~\ref{tab:supervised-results}), which we summarize by five per-seed scores,
we instead use a paired seed-level bootstrap against the BCE baseline over the
shared seeds. Full deltas and \(p\)-values for both families of tests are
tabulated in Appendix~\ref{app:significance}.

\paragraph{LLM diagnostic.}
The diagnostic runs Qwen2.5-72B-Instruct (4-bit quantization) with deterministic
decoding (temperature \(0\), at most 128 new tokens) under the two prompts of
Appendix~\ref{app:prompts}. Responses are parsed against the exact 19-label set;
unparseable responses are counted in the invalid-output rate and contribute no
labels.

\paragraph{Reproducibility.}
To preserve anonymity, we describe software and release plans in general terms.
Configurations, training and decoding scripts, tuned thresholds, and model
predictions will be released as a project artifact after review.\footnote{Code,
configurations, tuned thresholds, and model predictions
\ifpreprint are available at
\url{https://github.com/VictorMYeste/schwartz-geometry-value-detection}%
\fi.}

\section{Results}

\subsection{Training-Time Geometry Gives Limited Gains}

Table~\ref{tab:supervised-results} reports the supervised systems. BCE is a
strong, stable baseline (Macro-F1 \(0.2934\)).\footnote{For reference, prior work
on the same sentence-level task reports a comparable direct-classifier Macro-F1
(\(\approx\!0.281\), \texttt{deberta-base}, fixed \(0.5\) threshold)
\citep{yeste2026sentence}; the setups are not strictly comparable (we use
\texttt{deberta-v3-base} with tuned per-label thresholds), so we read this only as
a check that our baseline is competitive.} Asymmetric loss does not help
here: it is slightly weaker on every standard metric and much less stable across
seeds (Macro-F1 \(0.2833\!\pm\!0.0193\); circular error \(0.374\!\pm\!0.181\)).
The three GeoLoss-based variants (empirical, random, and Schwartz) all fall
within one standard deviation of BCE on the standard metrics, and the Schwartz
GeoLoss gain over BCE is not significant (paired bootstrap, \(p=0.19\) for
Macro-F1). They reduce expected circular error slightly (to \(0.127\)--\(0.131\)
from \(0.134\)), but the random control reduces it as much as the true geometry,
so the effect is not specific to the Schwartz structure. GeoSmooth is the clear
exception: in this formulation the soft targets destabilize training and collapse
performance (Macro-F1 \(0.1651\), significantly below BCE, \(p<0.001\)). These conclusions
are unchanged when the objectives are paired with the imbalance-aware ASL base
instead of BCE: all variants remain below their BCE counterparts and GeoSmooth
again collapses (Appendix~\ref{app:asl}). Training-time geometry thus yields no
consistent or theory-specific gain, motivating the post-hoc decoder.

\begin{table*}[t]
\centering
\small
\setlength{\tabcolsep}{5pt}
\begin{tabular}{lcccc}
\toprule
Method & Macro-F1\,$\uparrow$ & Micro-F1\,$\uparrow$ & Macro-AUPRC\,$\uparrow$ & Circ.\ err.\,$\downarrow$ \\
\midrule
BCE & 0.2934\,$\pm$\,0.0037 & 0.3425\,$\pm$\,0.0060 & 0.2353\,$\pm$\,0.0067 & 0.1342\,$\pm$\,0.0138 \\
ASL & 0.2833\,$\pm$\,0.0193 & 0.3306\,$\pm$\,0.0219 & 0.2235\,$\pm$\,0.0240 & 0.3735\,$\pm$\,0.1812 \\
Empirical structure & 0.2945\,$\pm$\,0.0053 & 0.3401\,$\pm$\,0.0048 & \textbf{0.2356}\,$\pm$\,0.0074 & \textbf{0.1269}\,$\pm$\,0.0101 \\
Random GeoLoss & 0.2949\,$\pm$\,0.0058 & \textbf{0.3439}\,$\pm$\,0.0085 & 0.2353\,$\pm$\,0.0059 & 0.1273\,$\pm$\,0.0105 \\
Schwartz GeoLoss & \textbf{0.2958}\,$\pm$\,0.0051 & 0.3421\,$\pm$\,0.0086 & \textbf{0.2356}\,$\pm$\,0.0051 & 0.1305\,$\pm$\,0.0096 \\
Schwartz GeoSmooth & 0.1651\,$\pm$\,0.0063 & 0.1656\,$\pm$\,0.0105 & 0.1182\,$\pm$\,0.0053 & 0.6988\,$\pm$\,0.0758 \\
\bottomrule
\end{tabular}
\caption{Supervised test results (mean\,$\pm$\,std, five seeds). $\uparrow$/$\downarrow$
= higher/lower is better. BCE and the three GeoLoss variants coincide within seed
noise on standard metrics; GeoSmooth collapses. Best per column \textbf{bold}.}
\label{tab:supervised-results}
\end{table*}

\subsection{Schwartz Decoding Improves Label-Set Coherence}
\label{sec:decoder-results}

The decoder is the main positive result. Applied to the BCE classifier
(Table~\ref{tab:decoder-main}), the Schwartz decoder leaves predictive performance
essentially unchanged---Macro-F1 \(0.2934\!\to\!0.2943\), Micro-F1
\(0.3425\!\to\!0.3430\), neither significant---while lowering the \emph{decoder
geometry cost} (the corpus-level sum of opposite-error rate, neighbor-error rate,
and confusion--distance correlation; Section~\ref{sec:setup}) from \(0.5634\) to
\(0.5480\). Two caveats keep this honest: the F1 preservation is guaranteed by the
Pareto selection rule (Section~\ref{sec:decoder}), not discovered, and the
reduction against the decoder's \emph{own} thresholding baseline is a sanity
check---the substantive evidence is the controlled comparison of
Section~\ref{sec:true-geometry}. The reduction is driven by the opposite-error rate
(\(0.5092\!\to\!0.5072\)) and the confusion--distance correlation
(\(-0.137\!\to\!-0.154\)); the neighbor-error rate rises slightly
(\(0.191\!\to\!0.195\); Section~\ref{sec:analysis}).

The edits are small and surgical: the decoder changes only \(2.45\%\!\pm\!0.26\%\)
of test sentences (\(\approx\!357\) of \(14{,}569\)), so the average set size barely
moves (\(0.804\!\to\!0.803\) labels per sentence; total predicted labels
\(11{,}718\!\to\!11{,}699\), \(\approx\!0.16\%\)). With the validation-selected
cardinality weight at \(\gamma=0\), the coherence gain is not an artifact of
predicting more or fewer labels. The claim is thus deliberately narrow: more
theory-coherent label sets at no measurable cost to F1.

\begin{table*}[t]
\centering
\small
\setlength{\tabcolsep}{5pt}
\begin{tabular}{lcccc}
\toprule
Decoder & Macro-F1\,$\uparrow$ & Micro-F1\,$\uparrow$ & Opp.\ err.\,$\downarrow$ & Geom.\ cost\,$\downarrow$ \\
\midrule
BCE thresholding & 0.2934\,$\pm$\,0.0037 & 0.3425\,$\pm$\,0.0060 & 0.5092\,$\pm$\,0.0096 & 0.5634\,$\pm$\,0.0120 \\
\quad + empirical decoder & 0.2935\,$\pm$\,0.0037 & 0.3428\,$\pm$\,0.0063 & 0.5100\,$\pm$\,0.0092 & 0.5628\,$\pm$\,0.0119 \\
\quad + random decoder & 0.2935\,$\pm$\,0.0038 & 0.3429\,$\pm$\,0.0062 & 0.5099\,$\pm$\,0.0091 & 0.5625\,$\pm$\,0.0115 \\
\quad + Schwartz decoder & \textbf{0.2943}\,$\pm$\,0.0034 & \textbf{0.3430}\,$\pm$\,0.0063 & \textbf{0.5072}\,$\pm$\,0.0098 & \textbf{0.5480}\,$\pm$\,0.0130 \\
\bottomrule
\end{tabular}
\caption{Structured decoder on the BCE classifier (mean\,$\pm$\,std, five seeds),
run under each geometry; Opp.\ err.\ and Geom.\ cost lower-is-better. F1 is preserved
everywhere, but only the Schwartz geometry materially lowers geometry cost. Best
\textbf{bold}.}
\label{tab:decoder-main}
\end{table*}

\subsection{The True Schwartz Geometry Matters}
\label{sec:true-geometry}

The decoder rows of Table~\ref{tab:decoder-main} already show that the random and
empirical geometries barely move the geometry cost; the cleanest test, however,
is a direct paired comparison. Relative to the controls run through the
identical decoder, the Schwartz geometry lowers the decoder geometry cost by
\(0.0145\) (vs.\ random; \(95\%\) bootstrap CI \([0.0074, 0.0231]\)) and
\(0.0148\) (vs.\ empirical; CI \([0.0074, 0.0231]\)), significant in all five
seeds (\(p<0.001\); the per-seed tests share the same direction, so we report
them without multiple-comparison correction), whereas the corresponding Macro-F1
and Micro-F1 differences are negligible (\(\le\!0.001\)) and significant in none. Matching the label
count, the circular form, or the empirical co-occurrence structure is therefore
not sufficient: the coherence gain comes from the Schwartz ordering itself.

\subsection{Prompted LLMs Do Not Replace Supervised Structured Prediction}

Table~\ref{tab:llm-diagnostic} compares the two prompted Qwen2.5-72B-Instruct
configurations with the supervised systems. Both prompts trail the supervised
models by a wide, significant margin on standard metrics: Qwen reaches Macro-F1
\(0.2430\)/\(0.2396\) and Micro-F1 \(0.2730\)/\(0.2643\) (definitions/continuum)
against \(0.2934\)/\(0.3425\) for BCE thresholding (5/5 seeds, \(p<0.001\) on
both). This is not a formatting artifact---the invalid-output rate is below
\(0.2\%\) under both prompts. Adding the Schwartz continuum to the prompt does
shift geometry-aware behavior, lowering the geometry cost from \(0.5757\) to
\(0.5633\) (\(p=0.01\)) and roughly doubling the average number of predicted labels
(\(1.07\!\to\!1.50\)), though it slightly lowers F1. The continuum-prompted cost
(\(0.5633\)) essentially matches untuned BCE thresholding (\(0.5634\)) but not the
decoder's (\(0.5480\)): prompting reaches the baseline's coherence, not the
decoder's. The comparison is intentionally asymmetric---zero-shot prompting against
a classifier fine-tuned on \(\approx\!45\)k in-domain sentences---so it is a bounded
probe of whether prompting suffices, and the supervised win is expected. Prompted
theory thus moves behavior in the expected direction without matching supervised
structured prediction.

\begin{table*}[t]
\centering
\small
\setlength{\tabcolsep}{6pt}
\begin{tabular}{lrrrr}
\toprule
System & Macro-F1\,$\uparrow$ & Micro-F1\,$\uparrow$ & Geom.\ cost\,$\downarrow$ & Inval.\ \%\,$\downarrow$ \\
\midrule
Qwen definitions & 0.2430 & 0.2730 & 0.5757 & \textbf{0.08} \\
Qwen continuum & 0.2396 & 0.2643 & 0.5633 & 0.14 \\
BCE thresholding & 0.2934 & 0.3425 & 0.5634 & --- \\
BCE + Schwartz decoder & \textbf{0.2943} & \textbf{0.3430} & \textbf{0.5480} & --- \\
\bottomrule
\end{tabular}
\caption{LLM diagnostic vs.\ supervised systems (LLM: single deterministic run;
supervised: five-seed mean). ``Inval.\ \%'': share of responses not parsable without
repair. The continuum prompt lowers geometry cost but not below the decoder, and
both prompts trail supervised F1. Best per column \textbf{bold}.}
\label{tab:llm-diagnostic}
\end{table*}

\section{Analysis and Discussion}
\label{sec:analysis}

A consistent pattern across our experiments is that the Schwartz geometry helps
as a post-hoc decoder but not when injected into the training loss. We read this
as a division of labor. The supervised encoder already learns strong local
evidence for each value, so a geometry penalty on the loss mostly perturbs an
already-good objective---and for GeoSmooth the cross-value soft targets, in this
particular formulation, are aggressive enough to destabilize training and collapse
performance. Independent
thresholding, by contrast, discards label dependencies at the final decision
step, which is exactly where the decoder operates: it reconciles per-label
evidence into a coherent set without altering the learned representations. The
same view explains why the training-time circular-error gains are small and not
theory-specific---a random ordering helps as much as the true one
(Table~\ref{tab:supervised-results})---while the decoder's neighbor and opposite
terms act on the discrete label set, where the ordering does matter.

The controlled comparison isolates this effect: only the true Schwartz ordering
lowers the decoder geometry cost, whereas a random circular permutation and an
empirical co-occurrence graph do not. Qualitatively, the decoder edits
predictions in the two ways the theory anticipates. It suppresses opposite-side
false positives---for a sentence whose gold label is \emph{Power: dominance},
thresholding also fires \emph{Benevolence: dependability} and
\emph{Universalism: concern}, two self-transcendence values on the far arc, which
the decoder removes---and it completes near-neighbor sets, as when thresholding
emits only \emph{Power: dominance} for a gold \emph{Power: resources} and the
decoder adds the adjacent true value. Both edits raise sample-level F1 while
reducing distant activations; Appendix~\ref{app:examples} tabulates these and
further examples.

The neighbor-error rate needs care, as it enters the minimized cost yet the decoder
completes neighbor sets. It counts only \emph{false-positive} neighbors (within two
steps of a gold value): completing a correct neighbor removes a near-miss and lowers
it, while adding a wrong one raises it. Because the decoder adds some unsupported
neighbors, the corpus rate rises slightly (\(0.191\!\to\!0.195\)) and the net gain
comes from the opposite-error and confusion-distance terms. This is why we minimize
the composite, not any single term, and read the controlled contrast as decisive.

We are deliberate about what improves. The decoder targets the structured label
set, not per-label calibration, so its gains surface in label-set coherence rather
than probability-mass metrics or large F1 jumps; Macro-F1 and Micro-F1 stay within
seed noise. The contribution is thus a more theory-consistent decision layer at no
measurable cost to accuracy, not a new state of the art---by design, since the
Schwartz circle is a soft bias and real texts express genuine value conflict that a
coherence objective should discourage only when unsupported, not forbid.

This soft, post-hoc use of the theory aligns with findings that hard architectural
encodings of Schwartz structure are brittle on this task
\citep{yeste2026hierarchical,yeste2026sentence}: our decoder leaves the flat
classifier intact and adds structure only as a final, tunable adjustment. The LLM
diagnostic points the same way from the other side: prompting the continuum steers
Qwen2.5-72B-Instruct toward more geometry-coherent behavior but trades away F1 and
stays well below the supervised systems. The controllable supervised decoder is thus
the more reliable way to make predictions respect the value space's structure.

\section{Limitations, Ethics, and Conclusion}

\paragraph{Limitations.}
Our scope is deliberately narrow: one dataset family (Touch\'e24-ValueEval),
English, sentence-level inputs, and one backbone (DeBERTa-v3-base); we do not test
other languages, domains, longer contexts, or larger encoders. The improvement we
measure is coherence, not accuracy---the decoder lowers theory-aware costs while
leaving Macro-F1 and Micro-F1 within seed noise---so its value is more consistent,
interpretable label sets, not higher standard scores. The decoder geometry cost is a
composite we define rather than an established benchmark, and the circular
operationalization (equal angular spacing, canonical order, opposite threshold
\(D>0.75\), two-step neighbor window) is one reasonable choice among several; a
mis-specified geometry could in principle suppress genuine value conflict, which the
soft penalty discourages but does not forbid. Because the decoder reranks candidate
sets from the classifier's own probabilities, it cannot recover values the base
model never surfaces and inherits any miscalibration of those probabilities and
thresholds. Finally, the LLM diagnostic uses a single model and two prompts under
deterministic decoding; it is a bounded probe whose numbers may shift with the
model, prompt, or decoding.

\paragraph{Ethics.}
Human value detection can support research on social, political, and moral
language, but it can also be misused to profile individuals or infer sensitive
beliefs. Sentence-level value attributions are uncertain and culturally
variable, and our systems should be read as tools for aggregate analysis and
annotation support, not as verdicts about individual speakers. Theory-aware
decoding improves structural consistency with the Schwartz taxonomy; it does not
remove annotation noise, ambiguity, or cross-cultural differences in how values
are expressed, and it should not be treated as evidence that a person holds a
value.

\paragraph{Conclusion.}
Human value detectors need not treat the refined Schwartz values as independent
labels. Injecting the circular structure into the training loss is informative but
not decisive---GeoLoss matches the baseline within noise and GeoSmooth degrades
it---whereas a post-hoc Schwartz-aware decoder, holding Macro-F1 and Micro-F1 fixed
by its selection rule, makes label sets measurably more coherent with the continuum,
and only when the geometry is the true one (random and empirical controls do not;
significant in all five seeds). Prompting an LLM with value definitions or the
continuum shifts behavior but does not close the gap to supervised structured
prediction. What sets this apart from prior work is where and how the theory
enters: unlike hard hierarchies or architectural gates on this task, which can
bottleneck recall \citep{yeste2026hierarchical,yeste2026sentence}, and unlike
structured decoders that learn label dependencies from data co-occurrence
\citep{read2011classifier,zhang2014review}, we fix the pairwise structure from
theory, keep it a soft bias rather than a hard constraint, and apply it post-hoc
to an untouched classifier, then test it directly against random and empirical
control geometries; to measure this faithfulness we also introduce a family of
theory-aware coherence metrics. More broadly, a psychological theory is most useful here as a soft,
controllable output-space bias applied at decoding time. We conjecture, but do not
test, that the recipe could extend to other label spaces with known scientific
structure (e.g.\ circumplex or wheel models of emotion); that transfer is left to
future work.

\ifpreprint
\section*{Acknowledgments}
The authors used Claude Opus 4.8 and Claude Fable 5 for language polishing,
structural editing, and assistance in drafting prose from author-provided notes,
tables, and verified experimental results. The authors reviewed and edited all
generated text and are responsible for all claims, analyses, and citations.
These models were also used to assist with code organization and
result-extraction scripts; all code and outputs were manually inspected by the
authors.
\fi

\bibliographystyle{acl_natbib}
\bibliography{tacl2021}

\appendix

\section{Geometry-Aware Training on the ASL Base}
\label{app:asl}

Table~\ref{tab:asl-geometry} repeats the geometry-aware objectives on the
imbalance-aware ASL base. The picture is unchanged: every variant trails the
BCE-based systems (Table~\ref{tab:supervised-results}), the GeoLoss variants do not
even beat the plain ASL baseline, and Schwartz GeoSmooth again collapses, more
severely than under BCE. Training-time geometry is thus not decisive regardless of
base loss, and the GeoSmooth instability is not specific to BCE.

\begin{table*}[tp]
\centering
\small
\setlength{\tabcolsep}{5pt}
\begin{tabular}{lcccc}
\toprule
Method (ASL base) & Macro-F1\,$\uparrow$ & Micro-F1\,$\uparrow$ & Macro-AUPRC\,$\uparrow$ & Circ.\ err.\,$\downarrow$ \\
\midrule
ASL (baseline) & 0.2833\,$\pm$\,0.0193 & 0.3306\,$\pm$\,0.0219 & 0.2235\,$\pm$\,0.0240 & 0.3735\,$\pm$\,0.1812 \\
\quad + empirical structure & \textbf{0.2879}\,$\pm$\,0.0048 & \textbf{0.3322}\,$\pm$\,0.0067 & \textbf{0.2306}\,$\pm$\,0.0042 & 0.1750\,$\pm$\,0.0167 \\
\quad + random GeoLoss & 0.2703\,$\pm$\,0.0040 & 0.3189\,$\pm$\,0.0028 & 0.2136\,$\pm$\,0.0044 & 0.1299\,$\pm$\,0.0058 \\
\quad + Schwartz GeoLoss & 0.2763\,$\pm$\,0.0054 & 0.3235\,$\pm$\,0.0090 & 0.2197\,$\pm$\,0.0050 & \textbf{0.1273}\,$\pm$\,0.0057 \\
\quad + Schwartz GeoSmooth & 0.2081\,$\pm$\,0.0146 & 0.2383\,$\pm$\,0.0173 & 0.1513\,$\pm$\,0.0107 & 0.8776\,$\pm$\,0.1307 \\
\bottomrule
\end{tabular}
\caption{Geometry-aware training on the ASL base (test, mean\,$\pm$\,std over
five seeds). All variants trail the BCE-based systems of
Table~\ref{tab:supervised-results}; the GeoLoss variants do not beat the plain
ASL baseline, and GeoSmooth collapses. F1 and AUPRC are higher-is-better;
circular error is lower-is-better. Best per column in \textbf{bold}.}
\label{tab:asl-geometry}
\end{table*}

\section{Per-Label Distribution}
\label{app:labels}

Table~\ref{tab:per-label} reports per-value support over the full corpus, in
canonical Schwartz order. Support spans roughly two orders of magnitude
(\emph{Humility} to \emph{Security: societal}), motivating the imbalance-aware
baseline (Section~\ref{sec:methods}) and validation-tuned per-label thresholds.

\begin{table}[t]
\centering
\small
\setlength{\tabcolsep}{4pt}
\begin{tabular}{lcr}
\toprule
Value & Region & Support (\%) \\
\midrule
Self-direction: thought & O & 1.24 \\
Self-direction: action & O & 3.52 \\
Stimulation & O & 2.64 \\
Hedonism & O/SE & 0.82 \\
Achievement & SE & 6.38 \\
Power: dominance & SE & 4.53 \\
Power: resources & SE & 5.07 \\
Face & SE/C & 1.83 \\
Security: personal & C & 2.07 \\
Security: societal & C & 8.64 \\
Tradition & C & 1.36 \\
Conformity: rules & C & 6.19 \\
Conformity: interpersonal & C & 1.35 \\
Humility & C/ST & 0.24 \\
Benevolence: dependability & ST & 1.95 \\
Benevolence: caring & ST & 2.28 \\
Universalism: concern & ST & 4.89 \\
Universalism: nature & ST & 2.15 \\
Universalism: tolerance & ST & 1.04 \\
\bottomrule
\end{tabular}
\caption{Per-label support as a percentage of all corpus sentences
(\(N=74{,}231\)), in canonical Schwartz order. Regions: O = openness to change,
SE = self-enhancement, C = conservation, ST = self-transcendence; Hedonism
(O/SE) and Face (SE/C) bridge two regions.}
\label{tab:per-label}
\end{table}

\section{Full Significance Tests}
\label{app:significance}

Table~\ref{tab:seed-significance} reports the paired seed-level bootstrap
(Section~\ref{sec:setup}) behind Table~\ref{tab:supervised-results}: the mean
test-set delta of each supervised variant against BCE over the five shared
seeds, with two-sided \(p\)-values. No GeoLoss variant differs significantly
from BCE on a standard metric; only the empirical control significantly
reduces circular error, and GeoSmooth is significantly worse on every metric.
Table~\ref{tab:control-significance} reports the paired sample-level bootstrap
behind Section~\ref{sec:true-geometry}: per-seed tests of the Schwartz decoder
against each control geometry on the shared test set. The geometry-cost
reduction is significant in every seed against both controls, while the F1
differences are significant in none.

\begin{table*}[tp]
\centering
\small
\setlength{\tabcolsep}{4pt}
\begin{tabular}{lrcrcrcrc}
\toprule
 & \multicolumn{2}{c}{Macro-F1} & \multicolumn{2}{c}{Micro-F1}
 & \multicolumn{2}{c}{Macro-AUPRC} & \multicolumn{2}{c}{Circ.\ err.} \\
\cmidrule(lr){2-3}\cmidrule(lr){4-5}\cmidrule(lr){6-7}\cmidrule(lr){8-9}
Method (vs.\ BCE) & \(\Delta\) & \(p\) & \(\Delta\) & \(p\) & \(\Delta\) & \(p\) & \(\Delta\) & \(p\) \\
\midrule
ASL & \(-0.0101\) & 0.170 & \(-0.0119\) & 0.193 & \(-0.0118\) & 0.170 & \(+0.2393\) & \textbf{\(<\)0.001} \\
Empirical structure & \(+0.0012\) & 0.721 & \(-0.0024\) & 0.186 & \(+0.0002\) & 0.878 & \(-0.0073\) & \textbf{0.008} \\
Random GeoLoss & \(+0.0015\) & 0.488 & \(+0.0014\) & 0.640 & \(-0.0001\) & 0.775 & \(-0.0069\) & 0.097 \\
Schwartz GeoLoss & \(+0.0025\) & 0.191 & \(-0.0004\) & 0.809 & \(+0.0003\) & 0.882 & \(-0.0037\) & 0.164 \\
Schwartz GeoSmooth & \(-0.1282\) & \textbf{\(<\)0.001} & \(-0.1769\) & \textbf{\(<\)0.001} & \(-0.1172\) & \textbf{\(<\)0.001} & \(+0.5646\) & \textbf{\(<\)0.001} \\
\bottomrule
\end{tabular}
\caption{Paired seed-level bootstrap for the supervised systems of
Table~\ref{tab:supervised-results}: mean test delta vs.\ BCE over the five
shared seeds (2{,}000 resamples; two-sided \(p\)). Positive \(\Delta\) favors
the variant on F1/AUPRC; negative \(\Delta\) favors it on circular error.
Significant \(p\) (\(<0.05\)) in bold.}
\label{tab:seed-significance}
\end{table*}

\begin{table}[tp]
\centering
\small
\setlength{\tabcolsep}{3pt}
\begin{tabular}{llrcc}
\toprule
Control & Metric & \(\Delta\) & Sig. & Max \(p\) \\
\midrule
Random & Geom.\ cost & \(-0.0145\) & 5/5 & \textbf{\(<\)0.001} \\
Random & Opp.\ err. & \(-0.0027\) & 4/5 & 0.96 \\
Random & Macro-F1 & \(+0.0008\) & 0/5 & 0.83 \\
Random & Micro-F1 & \(+0.0002\) & 0/5 & 0.77 \\
\midrule
Empirical & Geom.\ cost & \(-0.0148\) & 5/5 & \textbf{\(<\)0.001} \\
Empirical & Opp.\ err. & \(-0.0028\) & 4/5 & 0.96 \\
Empirical & Macro-F1 & \(+0.0008\) & 0/5 & 0.79 \\
Empirical & Micro-F1 & \(+0.0002\) & 0/5 & 0.80 \\
\bottomrule
\end{tabular}
\caption{Paired sample-level bootstrap for the decoder-control comparison of
Section~\ref{sec:true-geometry}, run per seed (2{,}000 resamples).
\(\Delta\) = Schwartz \(-\) control, averaged over seeds (negative is better
for cost and error); ``Sig.'' counts seeds with two-sided \(p<0.05\);
``Max \(p\)'' is the largest per-seed \(p\)-value, in bold when significant
(\(<0.05\)). The \(95\%\) confidence intervals for the geometry-cost deltas
appear in Section~\ref{sec:true-geometry}.}
\label{tab:control-significance}
\end{table}

\section{Qualitative Decoder Edits}
\label{app:examples}

Table~\ref{tab:decoder-examples} lists representative test sentences whose
label sets the Schwartz decoder edits, covering the two edit types discussed
in Section~\ref{sec:analysis}: completing a nearby true value and removing
opposite-side false positives (the third and fourth rows are the two examples
described there). Each row shows the gold labels, the thresholding output, the
decoded output, and the sentence-level F1 change.

\begin{table*}[tp]
\centering
\small
\setlength{\tabcolsep}{5pt}
\begin{tabular}{p{3.7cm}p{2.6cm}p{3.1cm}p{3.1cm}c}
\toprule
Sentence (abridged) & Gold & Thresholding & + Schwartz decoder & F1 \\
\midrule
\emph{\ldots fake advertisements and websites are designed with elements such
as discounts and vacation opportunities that consumers tend to be easily
persuaded\,\ldots} & Security: societal; Conformity: rules & Security:
personal & Security: personal; Security: societal; Conformity: rules &
0.00\,\(\to\)\,0.80 \\
\addlinespace
\emph{He stressed that the countries of the region should eventually join the
EU, of course within the framework of conditionality\,\ldots} & Conformity:
interpersonal & Self-direction: action; Conformity: rules & Conformity: rules;
Conformity: interpersonal & 0.00\,\(\to\)\,0.67 \\
\addlinespace
\emph{While in the field of regulatory reform it is possible to mark the
progress of the parties towards a compromise, when it comes to raising the
retirement age for women\,\ldots} & Power: dominance & Power: dominance;
Benevolence: dependability; Universalism: concern & Power: dominance &
0.50\,\(\to\)\,1.00 \\
\addlinespace
\emph{Provided, however, that the allied countries ``remain united in
maintaining and increasing sanctions pressure.''} & Power: resources & Power:
dominance & Power: dominance; Power: resources & 0.00\,\(\to\)\,0.67 \\
\bottomrule
\end{tabular}
\caption{Representative decoder edits on the test set, drawn from individual
seed runs. The first two rows complete nearby true values (the second also
removes an opposite-side false positive); the last two remove opposite-side
false positives or complete an adjacent value, matching the two examples
discussed in Section~\ref{sec:analysis}. F1 is the sentence-level F1 before
\(\to\) after decoding.}
\label{tab:decoder-examples}
\end{table*}

\section{LLM Diagnostic Prompts}
\label{app:prompts}

Both prompts are built from the fixed template below, reproduced verbatim from
the experiment code; the only per-sentence variation is the target sentence.
The \emph{continuum} prompt inserts the marked block between the definitions
and the output rules; the \emph{definitions} prompt omits it. Definitions
follow canonical Schwartz order. Italic parenthetical annotations are ours,
not part of the prompt.

\medskip
\begingroup\footnotesize\ttfamily\raggedright
\setlength{\parskip}{4pt}\setlength{\parindent}{0pt}
You are a sentence-level classifier for human values.

Task: identify which of the 19 refined Schwartz values are expressed in the
target sentence.

The labels collapse attained and constrained cases into value presence:
predict a label if the sentence expresses that value in either form.

Allowed labels and definitions:\\
- Self-direction: thought: freedom to cultivate one's own ideas and abilities.\\
- Self-direction: action: freedom to determine one's own actions.\\
- Stimulation: excitement, novelty, and change.\\
- Hedonism: pleasure and sensuous gratification.\\
- Achievement: success according to social standards.\\
- Power: dominance: power through exercising control over people.\\
- Power: resources: power through control of material and social resources.\\
- Face: maintaining one's public image and avoiding humiliation.\\
- Security: personal: safety in one's immediate environment.\\
- Security: societal: safety and stability in the wider society.\\
- Tradition: maintaining and preserving cultural, family, or religious
traditions.\\
- Conformity: rules: compliance with rules, laws, and formal obligations.\\
- Conformity: interpersonal: avoidance of upsetting or harming other people.\\
- Humility: recognising one's insignificance in the larger scheme of things.\\
- Benevolence: dependability: being a reliable and trustworthy member of the
in-group.\\
- Benevolence: caring: devotion to the welfare of in-group members.\\
- Universalism: concern: commitment to equality, justice, and protection for
all people.\\
- Universalism: nature: preservation of the natural environment.\\
- Universalism: tolerance: acceptance and understanding of those who are
different from oneself.

Schwartz-continuum structure: \emph{(continuum prompt only)}\\
- The refined values are arranged around this motivational circle:
Self-direction: thought -> Self-direction: action -> Stimulation ->
Hedonism -> Achievement -> Power: dominance -> Power: resources -> Face ->
Security: personal -> Security: societal -> Tradition -> Conformity: rules ->
Conformity: interpersonal -> Humility -> Benevolence: dependability ->
Benevolence: caring -> Universalism: concern -> Universalism: nature ->
Universalism: tolerance.\\
- Nearby labels on the circle usually express compatible motivations.\\
- Labels on the opposite side of the circle usually express motivational
conflict.\\
- Multi-label outputs are allowed, especially for nearby or conceptually
compatible values.\\
- Avoid predicting distant/opposing values together unless the sentence
clearly expresses both.

Output rules:\\
- Return exactly one JSON object with this schema:
\{"labels": ["Label name", "..."]\}.\\
- Use only exact labels from the allowed list. Do not invent labels.\\
- If no listed value is expressed, return \{"labels": []\}.\\
- Do not include explanations, markdown, comments, or text outside the JSON
object.

Target sentence:

[sentence] \emph{(the target sentence is substituted here)}
\endgroup

\end{document}